\documentclass[a4paper,11pt]{article}

\usepackage{geometry}
\usepackage[utf8]{inputenc}
\DeclareUnicodeCharacter{2212}{-}
\usepackage{amsmath}
\usepackage{microtype}
\usepackage{siunitx}
\usepackage{authblk}
\usepackage{lscape}
\usepackage{graphicx}
\usepackage{subcaption}
\usepackage[font={footnotesize},labelfont=bf]{caption}
\usepackage[super,numbers,sort&compress]{natbib}
\usepackage{booktabs}
\usepackage{tabularx}

\usepackage{blindtext}
\usepackage{booktabs,multirow,array}
\usepackage{tikz}
\usepackage{verbatim}
\usepackage{chemfig}
\usepackage{gensymb}
\usepackage{textcomp}
\usepackage{setspace}
\usepackage{multicol}
\usepackage{filecontents,pgfplots,pgfplotstable}
\usepackage{afterpage}
\usepackage{xtab,afterpage}
\usepackage{amssymb}
\usepackage{rotating}
\usepackage{natbib}
\usepackage{scrextend}
\usepackage{tcolorbox}
\usepackage[hidelinks]{hyperref}
\usepackage{multirow}
\usepackage{dirtree}
\usepackage{calc}
\usepackage{cleveref}
\usepackage{algorithm}
\usepackage[noend]{algpseudocode}
\usepackage{enumitem}
\setlist[itemize]{leftmargin=*}
\usepackage{microtype}
\usepackage{float}
\usepackage{forest}

{%

\usepackage{arxiv}
\bibliographystyle{naturemag}
\begin{document}
\title{MicroLib: A library of 3D microstructures generated from 2D micrographs using \textit{SliceGAN}}

\author[1]{{Steve Kench}
 \ }

\author[1]{{Isaac Squires}
 \ }
\author[1]{{Amir Dahari}
 \ }
\author[1]{Samuel J. Cooper \ \ }
\affil[1]{{\textit{\footnotesize Dyson School of Design Engineering, Imperial College London, London SW7 2DB}}}

\maketitle

\begin{abstract}
\begin{center}
\begin{minipage}{0.85\textwidth}
{\small 3D microstructural datasets are commonly used to define the geometrical domains used in finite element modelling. This has proven a useful tool for understanding how complex material systems behave under applied stresses, temperatures and chemical conditions. However, 3D imaging of materials is challenging for a number of reasons, including limited field of view, low resolution and difficult sample preparation. Recently, a machine learning method, \textit{SliceGAN}, was developed to statistically generate 3D microstructural datasets of arbitrary size using a single 2D input slice as training data. In this paper, we present the results from applying \textit{SliceGAN} to 87 different microstructures, ranging from biological materials to high-strength steels. To demonstrate the accuracy of the synthetic volumes created by \textit{SliceGAN}, we compare three microstructural properties between the 2D training data and 3D generations, which show good agreement. This new microstructure library both provides valuable 3D microstructures that can be used in models, and also demonstrates the broad applicability of the \textit{SliceGAN} algorithm.}
\end{minipage}
\end{center}
\end{abstract}
\vspace{.2cm}
\section*{Background}
\label{Background}
Understanding the influence of a material's microstructure on its performance has led to significant advancements in the field of material science \cite{gamble2011fabrication, plaut2007short, song2015differences}. Computational methods have played an important role in this success. For example, finite element analysis can capture complex stress fields during mechanical deformation of structural materials \cite{makarem2013nonlinear, ma2012finite}, and electro-chemical modelling can help to explain rate limiting factors during battery discharge \cite{wang2017finite, zadin2010modelling}. These simulations allow high-throughput exploration of a systems performance under a range of conditions \cite{liu2018microstructure, prabu2008microstructure}. In many fields, this has enabled massive acceleration of the materials optimisation process compared to experiments alone, and with significantly reduced cost. Importantly, 3D datasets are crucial for many applications where 2D datasets cannot be used to determine key material properties. For example, mechanical deformation, crack propagation and tortuosity are three material characteristics that behave fundamentally differently in 3D compared to 2D. 

The fidelity of the 3D microstructural datasets commonly required for physical modelling will influence the simulations reliability. Unfortunately, to the authors knowledge, there are no 3D material databases, with most data instead scattered across the literature. This is likely due to the high cost and technical experience required for 3D imaging techniques, which inhibits free sharing of data. Furthermore, where there is data available, it is commonly of limited resolution and field of view due to the intrinsic 3D imaging constraints of techniques such as focussed ion beam scanning electron microscopy and x-ray tomography \cite{cocco2013three}. In comparison, diverse, high resolution 2D micrographs are abundantly available online due to the prevalence of 2D imaging techniques such as light microscopy and scanning electron microscopy. DoITPoMS is one excellent micrograph repository with a broad range of alloys, ceramics, bio-materials and more \cite{doitpoms}. UHCSDB is a similar repository, focused solely on high carbon steels \cite{decost2017exploring}. ASM International has a collection of 4100 micrographs, though access costs a \$250 yearly subscription \cite{asminternational}.

In this paper, we aim to address the disparity between the availability of 2D micrographs compared to 3D. A number of previous approaches have been developed to address this problem through dimensionality expansion, which commonly entails statistical generation of 3D micrographs using statistics from a 2D training image. These are typically physic based and require the extraction of particular metrics from the training data for comparison. For example, sphere packing models using 2D particle size distributions \cite{jodrey1985computer}, poly-crystalline grain growth algorithms \cite{groeber2014dream}, and data fusion approaches \cite{oxinst}. 

In this work, we use \textit{SliceGAN}, a recently developed convolutional machine learning algorithm for dimensionality expansion\cite{kench2021generating}. A typical GAN uses two convolutional networks (generator and discriminator) to learn to mimic dataset distributions. The generator synthesises fake examples, and the discriminator identifies differences between these fake samples and the true training data distribution. Through iterative learning, the discriminator informs the generator how to make increasingly realistic samples that match the real training data. Importantly, in a typical set-up, the dimensionality of the generated images and the training data match. To facilitate different dimensionalities, \textit{SliceGAN} uses a simple modification; a 3D generator network produces a sample volume, then a 2D discriminator checks the fidelity of one slice at a time, where the 2D dimensionality of the slice now matches the 2D dimensionality of the training images. The algorithm is described in full in the original manuscript \cite{kench2021generating}. \textit{SliceGAN} is particularly well suited to the task at hand due to a number of key features. First, broad applicability means that the same algorithm and hyper-parameters can be used for a very diverse set of microstructures, as demonstrated in this dataset. Second, high speed training (typically 3 hrs on an RTX6000 GPU) and generation ($<3$ seconds for a 500$^3$ voxel volume) enables the synthesis of hundreds of large samples for statistical experiments, as well as the generation of volumes far larger than it is currently possible to obtain directly through imaging ($>2000^3$ voxel). Third, complete automation of the 2D to 3D algorithm is possible with no user defined inputs, such as statistical features, being required. This combination of strengths makes \textit{SliceGAN} an excellent candidate for building the first large scale 3D microstructural database from existing open-source 2D data.

\begin{figure}[h]
\centering
\includegraphics[width=\textwidth]{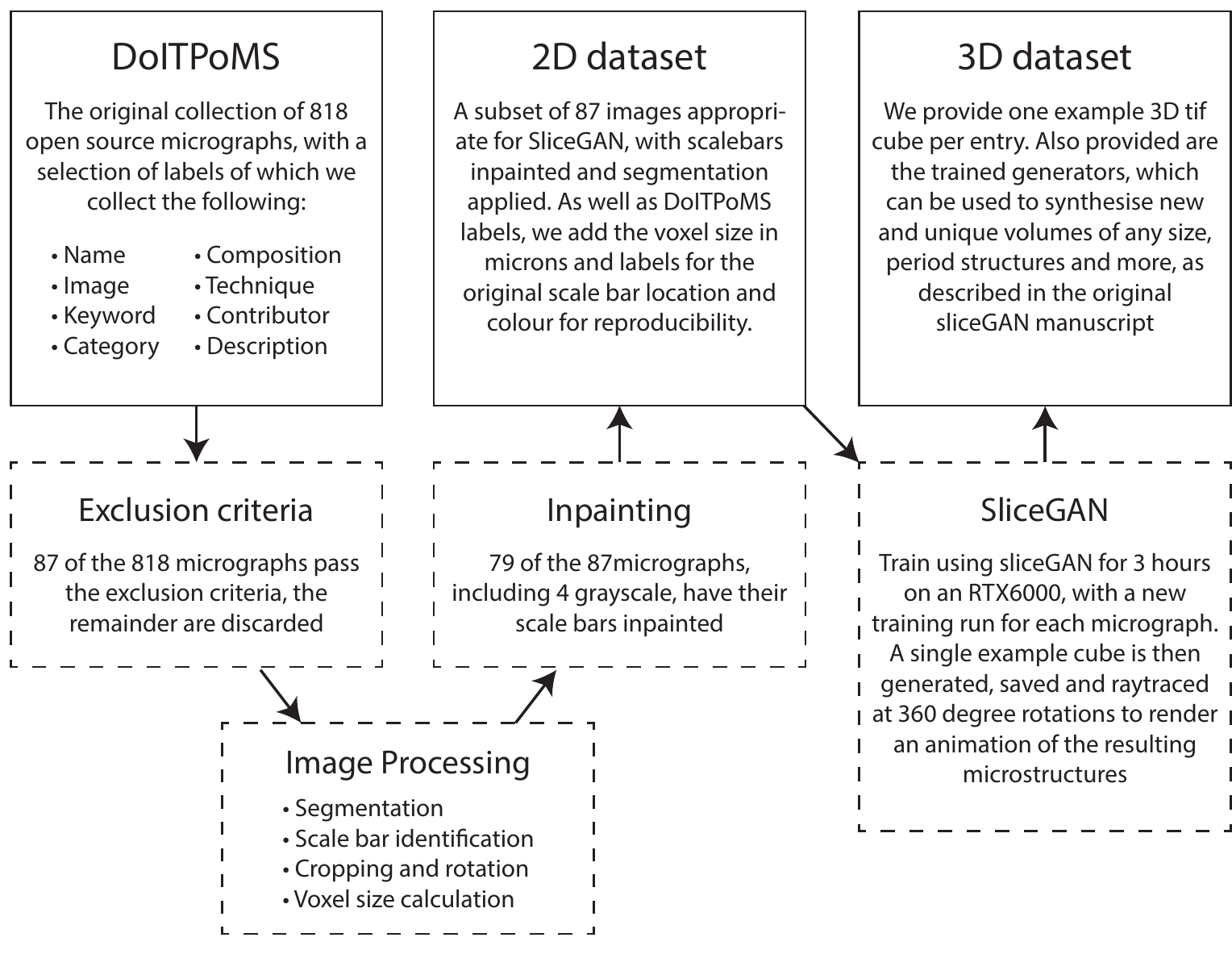}
\caption{{\footnotesize Starting from the DoITPoMS dataset, 4 key processing steps are used to generate the final 3D dataset, with an intermediate 2D cleaned and labelled dataset also available for download.}}
\label{F1}
\end{figure}

The benefits of this database are twofold. First, we provide a diverse 3D microstructural dataset which can be used by the material science community for modelling purposes. Crucially, users are not limited to the single example cube we provide, as each data entry also has an associated trained generator neural network (45 Mb in size) available to download. This can be used to synthesise arbitrary size datasets by cloning the \textit{SliceGAN} repo and running the relevant scripts (see methods). The second important function of this database is as a demonstration to the material science community of the strengths of \textit{SliceGAN}. The entries we provide are diverse in their nature, and contained in an easily searchable website. Interested researchers can thus use this website to check whether \textit{SliceGAN} works on materials in their research field, and see examples of generated outputs. This encourages the submission of more entries to the database, and the further use of \textit{SliceGAN} in the field of computational materials. The key data processing steps and datasets are presented in Figure \ref{F1}.

\section*{Methods}

As shown in Figure \ref{F1}, the database construction required several distinct steps. First, a subset of micrographs were selected using a set of exclusion criteria. A number of simple pre-processing operations were then applied to ensure suitability for the \textit{SliceGAN} workflow. An automated in-painting method was used to remove scale bars from the micrographs; compared to a cropping approach, this saves crucial data in an already extremely data-scarce setting. Finally, the resulting micrographs are used to train \textit{SliceGAN} generators, each of which was used to generate an example $320^3$ cubic volume. Each of these steps can be reproduced by cloning the \href{https://github.com/tldr-group/microlib}{MicroLib repository} and running \textit{main.py} in the relevant modes, as described in the repository \texttt{README}.

 \subsection*{Exclusion criteria}

DoITPoMS includes 818 diverse micrographs which can easily be downloaded directly from their website. However, not all are suitable for \textit{SliceGAN}, which has a number of limitations. As such, the following exclusion criteria are applied to leave 87 feasible microstructures:

\begin{enumerate}

\item Microstructure isotropy – \textit{SliceGAN} can be used for some anisotropic microstructures, but this mode requires multiple perpendicular micrographs which are not available from DoITPoMS.

\item Feature Representativity – \textit{SliceGAN} relies on feature statistics to generate realistic 3D volumes. Thus, a micrograph containing, for example, a single crystalline grain, is insufficient for the reconstruction of a 3D crystalline microstructure.

\item Even exposure – if parts of the micrograph are brighter than others, this creates significant issues in the final 3D volume, as the algorithm assumes homogeneity.

\item Uniqueness – In some cases there are several replicas of a similar microstructure; Different regions of the same material are always excluded, whilst where there are multiple magnifications, the max and min mag are used to capture different size features.

\item Image quality – Some micrographs are of too low quality to be worth reconstructing.

\end{enumerate}

\subsection*{Image processing}

This subsection of images were cropped to remove any borders or non-data regions, such as magnification information underneath the micrograph. Furthermore, of the 87 micrographs, 78 were identified as appropriate for segmentation as they contained easily distinguishable phases. Segmented images are preferable as most material simulation techniques require n-phase datasets such that phase properties can be assigned to a voxel. A simple thresholding process was applied to give n-phase micrographs, which are also better for the \textit{SliceGAN} training process due to their simplicity. The remaining 9 micrographs were processed as grayscale images. Finally, of the 87 microstructures, 79 had scale bars partially covering the micrograph. In these cases, the colour and location of the scale bar is identified and stored as an annotation to allow in-painting as described in the next section.

\subsection*{Scale bar inpainting}

Leaving scale bars in the training data images would result in \textit{SliceGAN} producing unrealistic features in the generated 3D volume, as it would interpret these objects as microstructural features. The simplest alternative is to entirely crop the region of the micrographs that contain the scale bar; however, this would result in a mean loss of 21\% of the training data (when then scale bar and label only actually conceal 1.4\% of the image on average). The quality of the final reconstructions could be significantly reduced due to a less representative and diverse distribution of features, which can lead to over-fitting and non-realistic microstructures. To avoid this scenario, we used a machine learning based in-painting technique to remove the scale bars, while leaving all surrounding data untouched. The locations of the scale bars are identified using a simple \textit{gui} which allows the user to select the scale bar colour and adjust a threshold until a sufficiently accurate mask is defined. A GAN is then trained to in-paint the masked image, as described in source. The resulting homogeneous microstructure is saved in the intermediate 2D dataset.

\newpage
\subsection*{\textit{SliceGAN} 3D reconstruction}

\begin{figure}[h]
\centering
\includegraphics[width=0.95\textwidth]{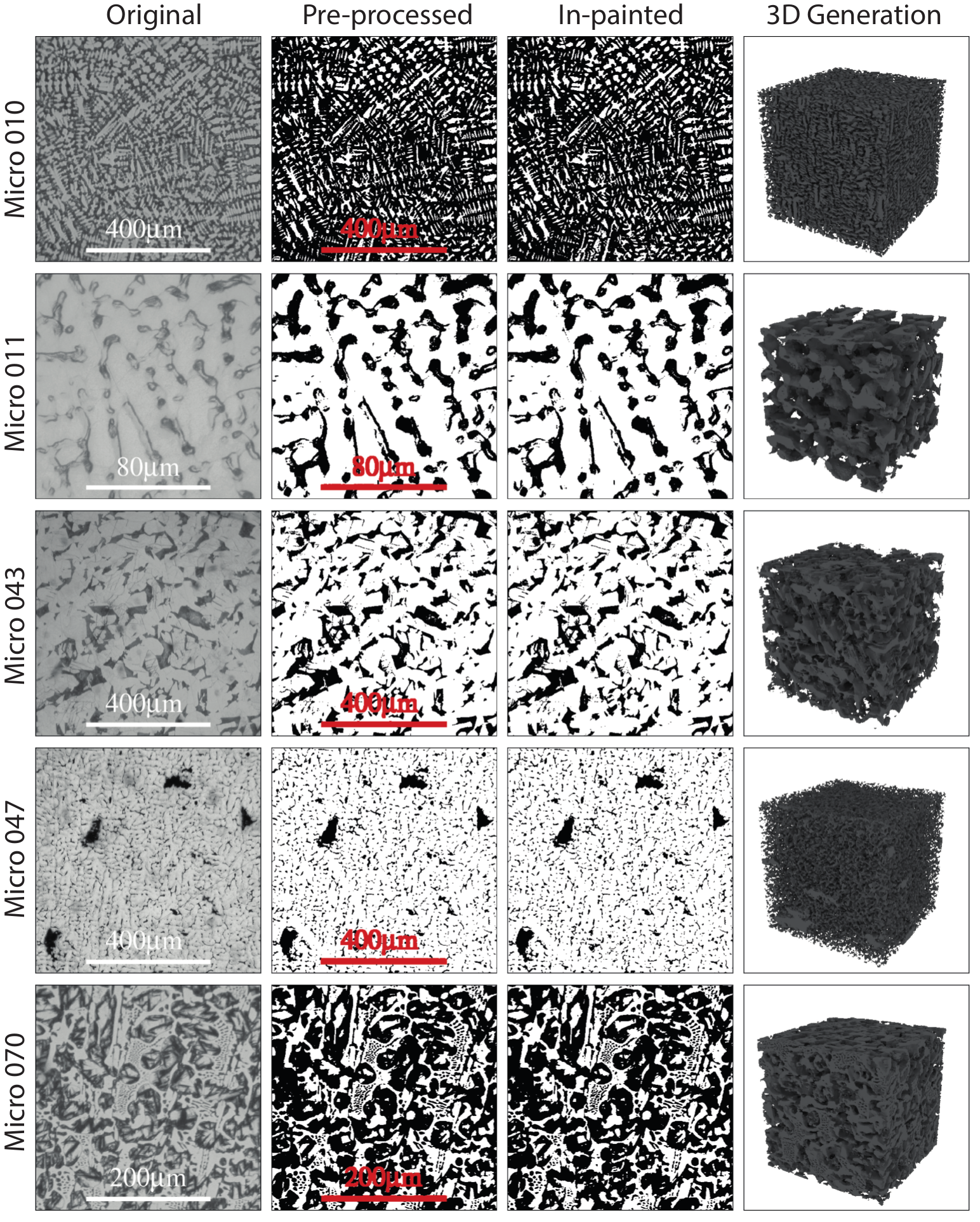}
\caption{{\footnotesize A selection of representative microstructures are shown, where each row depicts a micrograph, as well as each key stage of data processing required followed by the final 3D volume. Note the red masked scale bars in column 2 have been thickened by 1 pixel in each direction after thresholding of column 1. This ensures that all of the scale bar is in-painted, as edges are sometimes missed due to pixelization.}}
\label{F2}
\end{figure}

Each microstructure is trained on randomly initialised \textit{SliceGAN} networks for 5000 generator iterations, each with batch size 32, which takes less than 3 hours per entry using an RTX6000 GPU and x CPU. Hyper-parameters are kept consistent with the original \textit{SliceGAN} paper. The resulting trained generators are saved and then used to synthesise a 320 voxel cube, which takes less than 3 seconds. Figure \ref{F2} depicts the outputs at each step for a selection of microstructures.

\section*{Data Record}
The full dataset can be accessed at zenodo (\hyperlink{https://doi.org/10.5281/zenodo.7118559}{https://doi.org/10.5281/zenodo.7118559
} \cite{fs}). The dataset is compressed into a single zip file, which contains one sub-folder for each of the 87 entries selected from DoITPoMs. Sub-folder names are derived from the DoITPoMS microstructure ID. Each contains media files (.png, .gif, .mp4) that illustrate the steps taken to generate the data and display examples of resulting microstructures. They also contain trained models and a parameters file (.pt, .data) to allow users to synthesise new volumes.

As well as this set of sub-folders, the root directory also contains an annotations file, data\_anns.json. This json contains web-scraped descriptors of each microstructure (with the exception of data\_type, which was defined during this study). The full directory tree and brief descriptions of each file are given below.

\vspace{0.5cm}
\renewcommand{\F}[1]{\hspace{1cm}\textrm{#1}}
\newcommand{\Q}[2]{\parbox{0.43\textwidth}{#1}\textrm{#2}}
\newcommand{\T}[2]{\parbox{0.4\textwidth}{#1}\textrm{#2}}

\dirtree{%
.1 microlibDataset.
.2 microstructure001.
.3 \Q{microstructure001.mp4}{360 degree rotation of example volume}. 
.3 \Q{microstructure001.tif}{Example cube}. 
.3 \Q{microstructure001\_Disc.pt}{Trained discriminator parameters}.
.3 \Q{microstructure001\_Gen.pt}{Trained generator parameters}.
.3 \Q{microstructure001\_inpaint.mp4}{Movie showing inpaint process}.
.3 \Q{microstructure001\_inpaint\_gif.gif}{Gif showing inpaint process}.
.3 \Q{microstructure001\_inpainted.png}{Inpainted image used for training}.
.3 \Q{microstructure001\_long.png}{Extended rotation and erosion of example volume}.
.3 \Q{microstructure001\_original.png}{*Original raw image}.
.3 \Q{microstructure001\_params.data}{Parameters for loading trained models}.
.2 microstructure002.
.3 ....
.2 ....
.2 data\_anns.json.
.3 1.
.4 \T{name: microstructure001}{ID}.
.4 \T{data\_type: two-phase}{Two-phase or grayscale}.
.4 \T{brief\_descr: hyper-eutectoid}{DoITPoMS description}.
.4 \T{keywords: Aluminium, ...}{Searchable identifiers}.
.4 \T{category: Metal, Alloy}{Material type}.
.4 \T{element: Al, Cu}{List of elements in micrograph}.
.4 \T{technique: light microscopy}{Characterisation method}.
.4 \T{long desc: Micrograph shows...}{Full description}.
.4 \T{contributor: Prof T W Clyne}{Researcher who added sample}.
.4 \T{organisation: Dept. Mat Sci}{Org. of contributor}.
.3 2.
.4 ....
.3 ....
}

\begin{figure}[h]
\centering
\includegraphics[width=\textwidth]{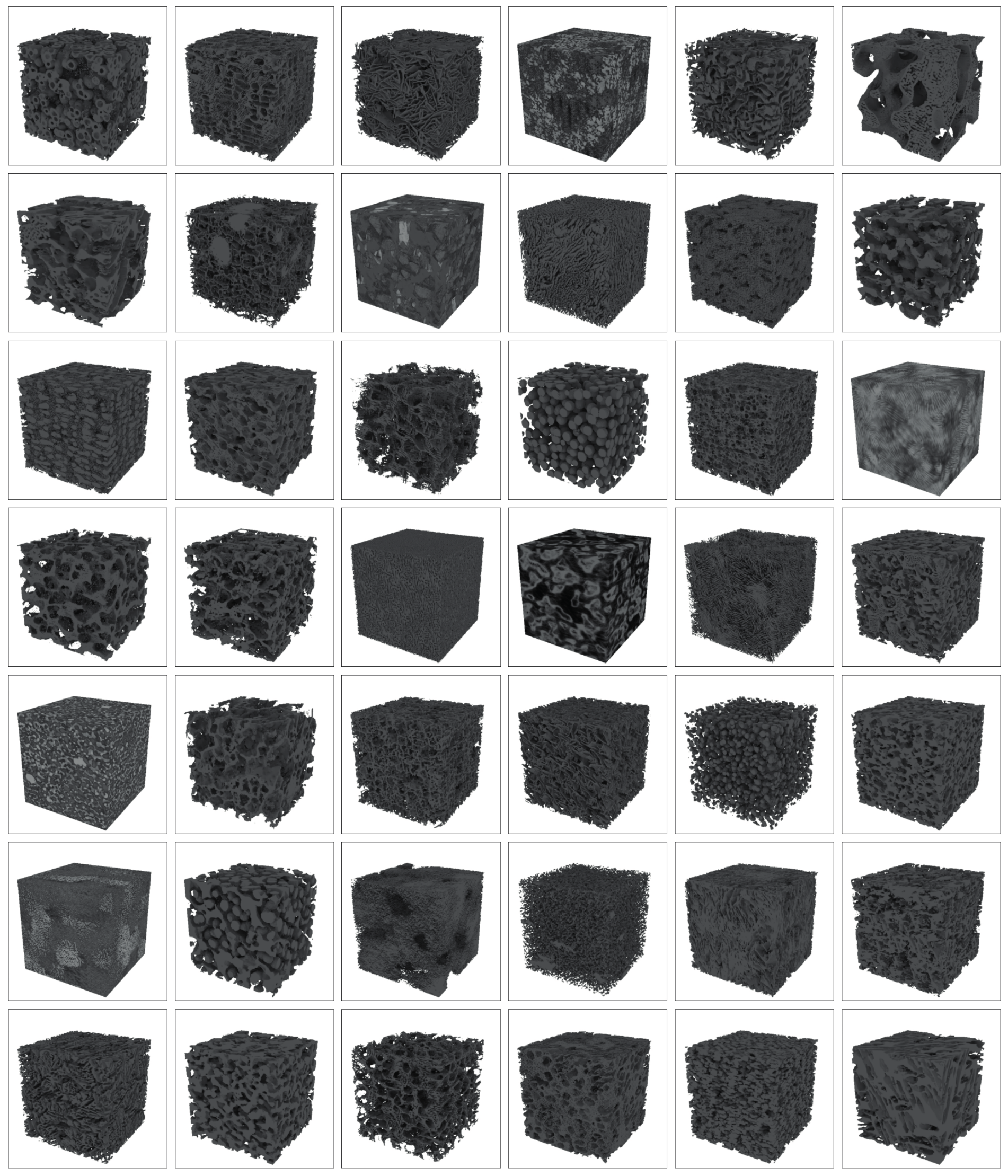}
\caption{{\footnotesize 42 of the 87 microstructures are depicted, selected based on their uniqueness rating. Each volume is $320^3$ voxels.}}
\label{F3}
\end{figure}

Besides zenodo, we have also created \hyperlink{https://microlilb.io}{microlib.io}, a website where users can visualise microstructures and search through the microlib database using keywords and filters. Individual media and model files can also be directly downloaded. The web app also provides an easy to use \href{https://microlilb.io/api}{API} which the user can query using the same search functionality as the website, allowing for programmatic access to the same data and metadata provided by the frontend. Figure \ref{F3} shows a selection of the 3D tifs available, though readers are encouraged to view the MicroLib website for the best visualisation experience.
\newpage
\section*{Technical validation}

\begin{figure}[h]
\centering

\includegraphics[width=\textwidth]{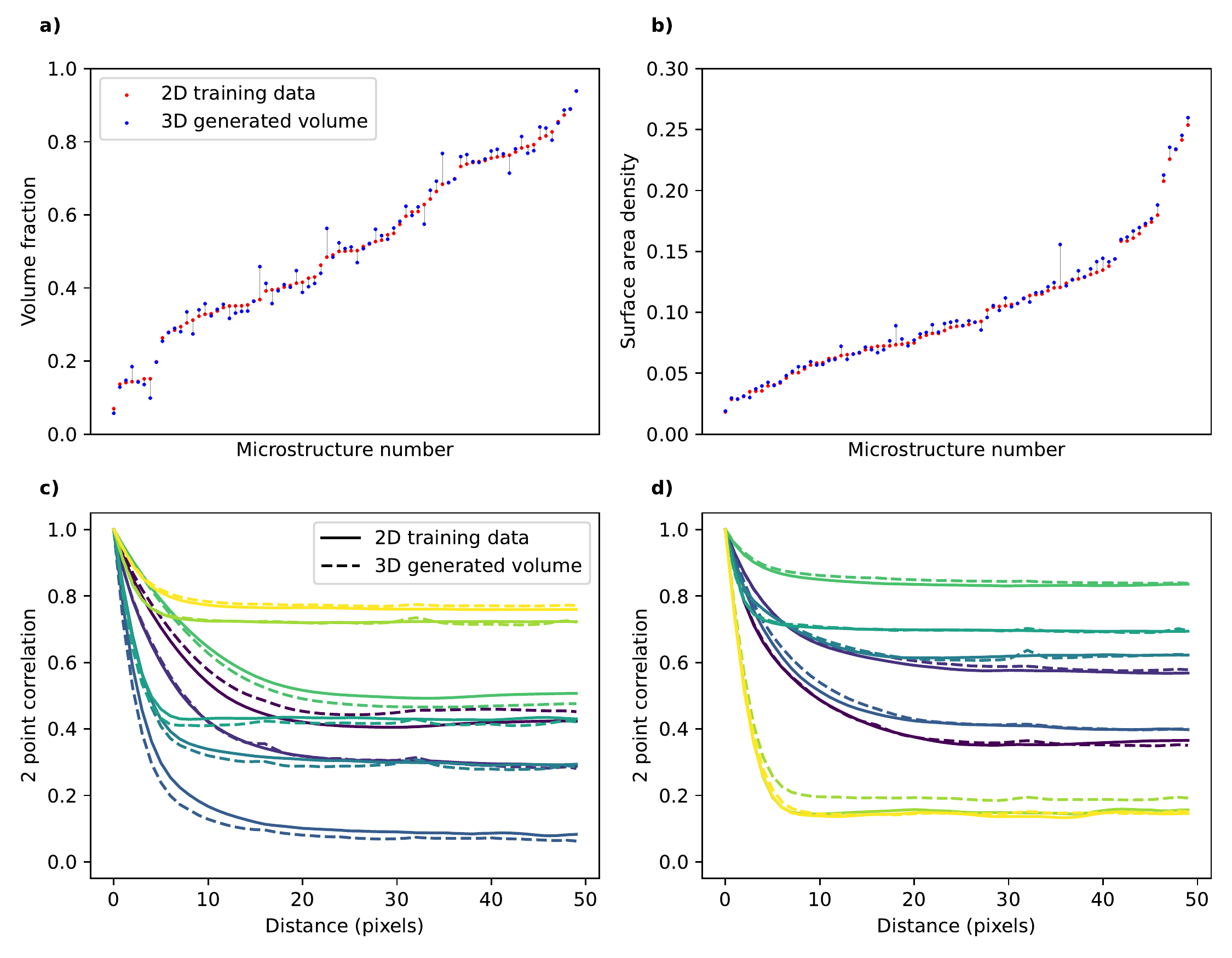}
\caption{{\footnotesize Property comparison for 2D training data vs 3D generated volumes. Plots are ordered by increasing value in the 2D dataset (volume fraction and surface area for a) and b) respectively. Plots b) and c) show the two-point correlation function for four randomly selected microstructures each. Within each plot, only lines of the same colour should be compared as they are from the same material.}}
\label{F4}
\end{figure}

Unlike some machine learning methods, such as auto-encoders, GANs do not attempt to exactly recreate images from the training set. Instead, they capture the underlying probability distribution of the training data and synthesise samples with the same distribution of features. This means that there is no ground truth against which the generated outputs can be compared. Thus, to quantify the accuracy of the generated 3D volumes compared to their original 2D training data, we calculate and compare a number of statistical material properties. These tests are only performed on the 78 n-phase materials, as the properties cannot be calculated for grayscale images. First, volume fraction (vf), which is simply the proportion of the voxels assigned to a particular phase. As shown in Figure \ref{F4}a), the 3D agrees well with the 2D training data; the mean percentage error, calculated as $\frac{\lvert\text{vf(2D)} - \text{vf(3D)}\rvert}{\text{vf(2D)}}$, is 4.7\%. Figure \ref{F4}b) shows a similar comparison for normalised surface area density, which is calculated as the proportion of voxel faces touching both phase 1 and phase 2. The mean percentage error is 4.3\%. 

As well as these simple metrics, we also can compare the two-point  correlation function (2PC) of the training data and generated volumes. The 2PC gives the probability of finding the same phase pixel at a given pixel separation distance, as describe in more detail in the original SliceGAN paper\cite{kench2021generating}. Figure \ref{F4}c and \ref{F4}d each show the 2PC of 8 randomly selected microstructure entries. In general, the curves are very similar for 2D and 3D.

Although the majority of samples show excellent agreement between 2D and 3D, there are a number of outliers, in particular for volume fraction. 5 samples (tags 1, 60, 372, 612, 782) show a volume fraction error greater than 5\% (Supplementary Figure S1 shows three of these microstructures, which exemplify the key failure modes to be discussed below). Notably, sample 60 is also the outlier in the surface area density plot. To explore the nature of this error, three repeats were run on the microstructure to test whether the generator would reproduce the observed behaviour. The new generators gave the same metrics to within 1\%. This implies that in this particular use case, well trained generators produce higher volume fractions in 3D than in 2D. 

Observing the microstructure itself gives some indication of why this might occur. Sample 60 consists mostly of ovals with a few circles, all of similar sizes. Under the assumption of isotropy, we can ask what 3D structure we expect \textit{SliceGAN} to generate, and indeed we quickly conclude there is no feasible isotropic 3D volume that can be made where all 2D slices contain only these features. Crucially, we are missing smaller ovals or circles that would be present at the edges of spherical or ellipsoidal features. \textit{SliceGAN} is thus forced to compromise between accurately reproducing volume fraction versus the exact feature distribution, as both are not possible. Sample 782 suffers from the same problem, whilst sample 1 and 372 have large non-representative features which lead to a similar scenario. Finally, sample 612 is simply poorly segmented. This demonstrates that poor agreement of metrics is one indicator that a non-representative, anisotropic or low quality 2D microstructure has been used to train \textit{SliceGAN}, which is potentially useful in catching cases where the exclusion criteria were insufficient. However, it is worth noting that unrealistic features might still occur even when volume fractions and two-point correlations match the 2D dataset well. As such, great care should be taken when using these results, and where possible, users of \textit{SliceGAN} should always compare 2D and 3D statistics for properties relevant to the simulations they are conducting.

\section*{Code availability}
All code for generating the datasets, including image scraping, preprocessing, inpainting and \textit{SliceGAN}, can be accessed openly at the MicroLib github repository, \href{https://github.com/tldr-group/microlib}{https://github.com/tldr-group/microlib}, which includes in-depth instruction to ensure reproducibility.

\section*{Acknowledgements}
This work was supported by funding from the EPSRC Faraday Institution Multi-Scale Modelling project
(https://faraday.ac.uk/; EP/S003053/1, grant number
FIRG003 received by SK).
The authors would also like to thank the authors of DoITPoMS for their open source dataset, without which this work would not be possible.

\section*{Author contributions}
SK developed the code required for the generation of the 2D dataset, performed the technical validation and wrote the paper. AD ran \textit{SliceGAN} scripts on the Imperial College high-performance to generate 3D volumes. IS developed the microlib.io website. All authors contributed to the conceptualisation of the project and editing of the paper.

\section*{Competing interests}
The authors declare no competing interests.

\addcontentsline{toc}{section}{References}


\end{document}


\title{Supplementary Information}
\author{}

\affil{}
\date{}
\maketitle
	
\section{Microstructures of interest}
\vspace{0cm}
\begin{figure}[h]
\centering

\includegraphics[width=0.8\textwidth]{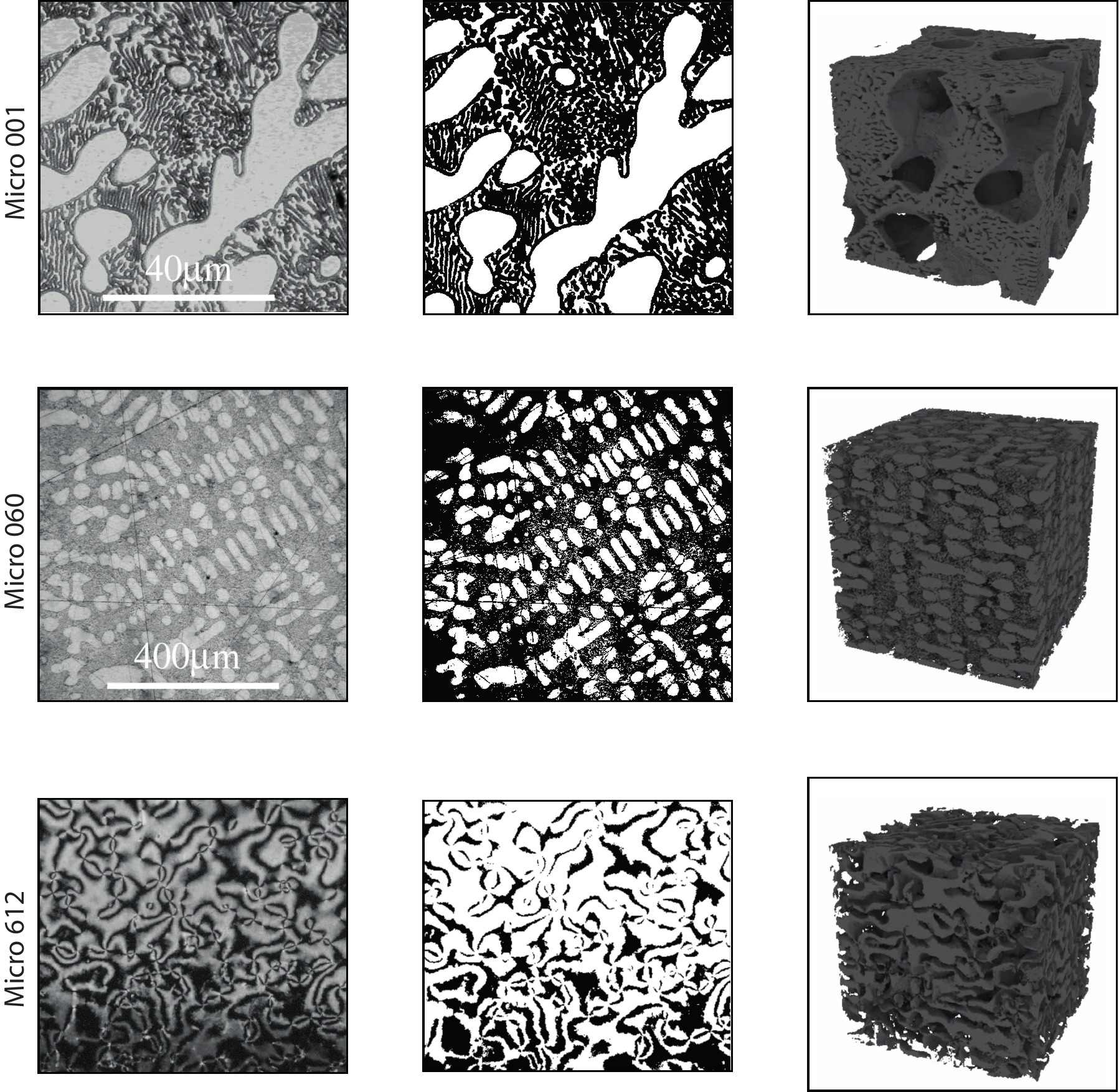}
\caption{{\footnotesize Examples showing potential modes of failure in \textit{SliceGAN}. Micro 1 has large white features, which are not unlikely to be statistically representative of the microstructure as a whole. Micro 60 has a large number of ovals, which cannot be made into 3D features without introducing new features such as smaller circles. Micro 612 has a graded brightness, resulting in poor segmentation.}}
\label{S1}
\end{figure}